# Analysis, Interpretation, and Recognition of Facial Action Units and Expressions Using Neuro-Fuzzy Modeling

Mahmoud Khademi[1], Mohammad Hadi Kiapour[2], Mohammad T. Manzuri-Shalmani[1], and Ali A. Kiaei[1]

[1] DSP Lab, Sharif University of Technology, Tehran, Iran
[2] Institute for Studies in Fundamental Sciences (IPM), Tehran, Iran
`{khademi@ce.,kiapour@ee.,manzuri@,kiaei@ce.}sharif.edu`

**Abstract.** In this paper an accurate real-time sequence-based system for representation, recognition, interpretation, and analysis of the facial action units (AUs) and expressions is presented. Our system has the following characteristics: 1) employing adaptive-network-based fuzzy inference systems (ANFIS) and temporal information, we developed a classification scheme based on neuro-fuzzy modeling of the AU intensity, which is robust to intensity variations, 2) using both geometric and appearance-based features, and applying efficient dimension reduction techniques, our system is robust to illumination changes and it can represent the subtle changes as well as temporal information involved in formation of the facial expressions, and 3) by continuous values of intensity and employing top-down hierarchical rule-based classifiers, we can develop accurate human-interpretable AU-to-expression converters. Extensive experiments on Cohn-Kanade database show the superiority of the proposed method, in comparison with support vector machines, hidden Markov models, and neural network classifiers.

## 1 Introduction

Human face-to-face communication is a standard of perfection for developing a natural, robust, effective and flexible multi modal/media human-computer interface due to multimodality and multiplicity of its communication channels. In this type of communication, the facial expressions constitute the main modality [1]. In this regard, automatic facial expression analysis can use the facial signals as a new modality and it causes the interaction between human and computer more robust and flexible. Moreover, automatic facial expression analysis can be used in other areas such as lie detection, neurology, intelligent environments and clinical psychology.

Facial expression analysis includes both measurement of facial motion (e.g. mouth stretch or outer brow raiser) and recognition of expression (e.g. surprise or anger). Real-time fully automatic facial expression analysis is a challenging complex topic in computer vision due to pose variations, illumination variations, different age, gender, ethnicity, facial hair, occlusion, head motions, and lower intensity of expressions. Two survey papers summarized the work of facial expression analysis before year 1999 [2, 3]. Regardless of the face detection stage, a typical automatic facial expression analysis system consists of facial expression data extraction and facial expression classification stages. Facial feature processing may happen either holistically, where the face is processed as a whole, or locally. Holistic feature extraction methods are good at determining prevalent facial expressions, whereas local methods are able to detect subtle changes in small areas.

There are mainly two approaches for facial data extraction: geometric-based methods and appearance-based methods. The geometric facial features present the shape and locations of facial components. With appearance-based methods, image filters, e.g. Gabor wavelets, are applied to either the whole face or specific regions in a face image to extract a feature vector [4].

The sequence-based recognition method uses the temporal information of the sequences (typically from natural face towards the frame with maximum intensity) to recognize the expressions. To use the temporal information, the techniques such as hidden Markov models (HMMs) [5], recurrent neural networks [6] and rule-based classifier [7] were applied.

The facial action coding system (FACS) is a system developed by Ekman and Friesen [8] to detect subtle changes in facial features. The FACS is composed of 44 facial action units (AUs). 30 AUs of them are related to movement of a specific set of facial muscles: 12 for upper face (e.g. AU 1 inner brow raiser, AU 2 outer brow raiser, AU 4 brow lowerer, AU 5 upper lid raiser, AU 6 cheek raiser, AU 7 lid tightener) and 18 for lower face (e.g. AU 9 nose wrinkle, AU 10 upper lip raiser, AU 12 lip corner puller, AU 15 lip corner depressor, AU 17 chin raiser, AU 20 lip stretcher, AU 23 lip tightener, AU 24 lip pressor, AU 25 lips part, AU 26 jaw drop, AU 27 mouth stretch). Facial action units can occur in combinations and vary in intensity. Although the number of single action units is relatively small, more than 7000 different AU combinations have been observed. To capture such subtlety of human emotion paralinguistic communication, automated recognition of fine-grained changes in facial expression is required (for more details see [8, 9]).

The main goal of this paper is developing an accurate real-time sequence-based system for representation, recognition, interpretation, and analysis of the facial action units (AUs) and expressions. We summarize the advantages of our system as follows:

1) The facial action unit intensity is intrinsically fuzzy. We developed a classification scheme based on neuro-fuzzy modeling of the AU intensity, which is robust to intensity variations. Applying this accurate method, we can recognize lower intensity and combinations of AUs.
2) Recent work suggests that spontaneous and deliberate facial expressions may be discriminated in term of timing parameters. Employing temporal information instead of using only the last frame, we can represent these parameters properly.
3) By using both geometric and appearance features, we can increase the recognition rate and also make the system robust against illumination changes.
4) By employing top-down hierarchical rule-based classifiers such as J48, we can automatically extract human interpretable classification rules to interpret each expression.
5) Due to the relatively low computational cost, the proposed system is suitable for real-time applications.

The rest of the paper has been organized as follows: In section 2, we describe the approach which is used for facial data extraction and representation using both geometric and appearance features. Then, we discuss the proposed scheme for recognition of facial action units and expressions in section 3 and section 4 respectively. Section 5 reports our experimental results, and section 6 presents conclusions and a discussion.

## 2 Facial Data Extraction and Representation

### 2.1 Biased Learning

Biased learning is a learning problem in which there are an unknown number of classes but we are only interested in one class. This class is called "positive" class. Other samples are considered as "negative" samples. In fact, these samples can come from an uncertain number of classes. Suppose $\{x_i | i = 1, ..., N_x\}$ and $\{y_i | i = 1, ..., N_y\}$ are the set of positive and negative d-dimensional samples (feature vectors) respectively. Consider the problem of finding d × r transformation matrix w (r ≪ d), such that separates projected positive samples from projected negatives in the new subspace. The dimension reduction methods like fisher discriminant analysis (FDA) and multiple discriminant analysis have addressed this problem simply as a two-class classification problem with symmetric treatment on positive and negative examples. For example in FDA, the goal is to find a subspace in which the ratio of between-class scatter over within-class scatter matrices is maximized. However, it is part of the objective function that negative samples shall cluster in the discriminative subspace. This is an unnecessary and potentially damaging requirement because very likely the negative samples belong to multiple classes. In fact, any constraint put on negative samples other than stay away from the positives is unnecessary and misleading. With asymmetric treatment toward the positive samples, Zhou and Huang [10] proposed the following objective function:

$$w_{opt} = \arg\max_W \frac{\text{trace}(w^T S_y w)}{\text{trace}(w^T S_x w)} \qquad (1)$$

where $S_y$ and $S_x$ are within-class scatter matrices of negative and positive samples with respect to positive centroid, respectively. The goal is to find w that clusters only positive samples while keeping negatives away. The problem of finding optimal w, becomes finding the generalized eigenvectors α's associated with the largest eigenvalues λ's in the below generalized eigenanalysis problem:

$$S_y \alpha = \lambda S_x \alpha \qquad (2)$$

Our goal is developing a facial action unit recognition system that can detect whether the AUs occur or not. The input of the system is a sequence of frames from natural face towards one of the facial expressions with maximum intensity. Suppose we have extracted a feature matrix or a feature vector from each frame. In order to embed facial features in a low-dimensionality space and deal with curse of dimensionality dilemma, we should use a dimension reduction method. For recognition of each AU, we are facing an asymmetric two-class classification problem. For example when the goal is detecting whether AU 27 (mouth stretch) occur or not, the positive class includes all of sequences in the train set that represent stretching of the mouth; other sequences are considered as negative samples. These samples can come from an uncertain number of classes. They can represent any single AU or AU combinations except AU 27. In fact, our problem is a biased learning problem.

### 2.2 Appearance-Based Facial Feature Extraction Using Gabor Wavelets

In order to extract the appearance-based facial features from each frame, we use a set of Gabor wavelets. They allow detecting line endings and edge borders of each frame over multiple scales and with different orientations. Gabor wavelets remove also most of the variability in images that occur due to lighting changes [4]. Each frame is convolved with p wavelets to form the Gabor representation of the t frames (Fig. 1).

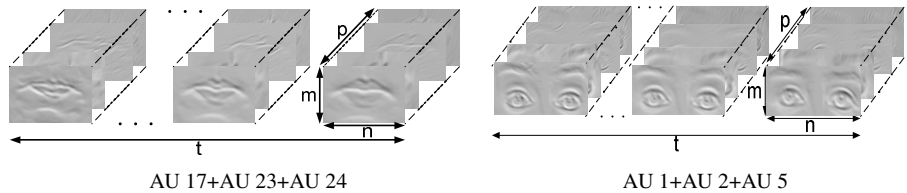

AU 17+AU 23+AU 24             AU 1+AU 2+AU 5

**Fig. 1.** Examples of the image sequences and their representation using Gabor wavelets

However, for applying the Zhou and Huang's method, which is called biased discriminant analysis (BDA), to the facial action unit recognition problem we should first transform the feature matrices of the sequence into a one-dimensional vector that ignores the underlying data structure (temporal and local information) and leads to the curse of dimensionality dilemma and the small sample size problem. Thus, we use two-dimension version of BDA algorithm by simply replacing the image vector with image matrix in computing the corresponding variance matrices to reduce the dimensionality of each feature matrix in two directions [11]. Then, we apply BDA algorithm to the vectorized representation of the reduced feature matrices. Also, In order to deal with singularity in the matrices we use 2D and 1D principle component analysis (PCA) algorithms [12], before applying 2DBDA and BDA respectively.

### 2.3 Geometric-Based Facial Feature Extraction Using Optical Flow

In order to extract geometric features we use a facial feature extraction method presented in [13]. The points of a 113-point grid, which is called Wincanide-3, are placed on the first frame manually. Automatic registering of the grid with the face has been addressed in many literatures (e.g. see [14]). For upper face and lower face action units a particular set of points are selected. The pyramidal optical flow tracker [15] is employed to track the points of the model in the successive frames towards the last frame (see Fig. 2). The loss of the tracked points is handled through a model deformation procedure (for detail see [13]). For each frame, the displacements of the points in two directions with respect to the first frame are calculated and placed in the columns of a matrix. Then, we apply 2DBDA algorithm [11] to the matrix in two directions. The vectorized representation of the reduced feature matrix is used as geometric feature vector.

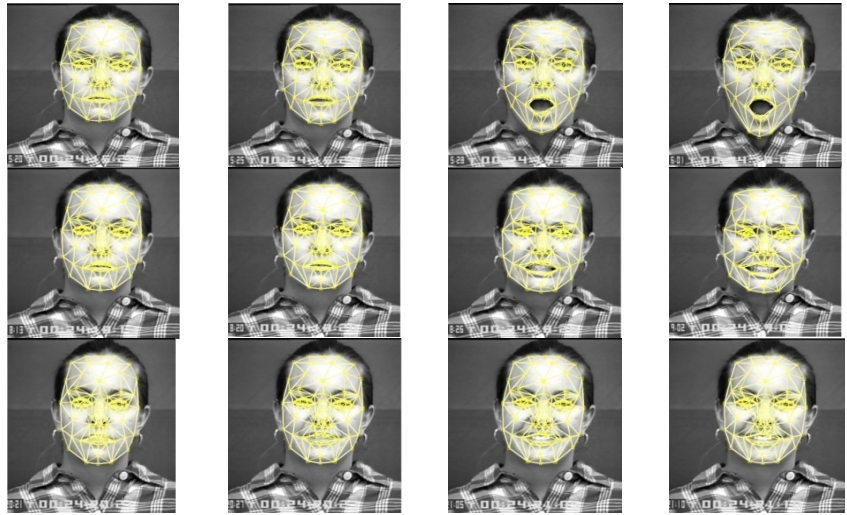

**Fig. 2.** Geometric-based facial feature extraction using grid tracking

## 3   Facial Action Unit Recognition Using Neuro-Fuzzy Modeling

### 3.1   Takagi-Sugeno Fuzzy Inference System and Training Data Set

The flow diagram of the proposed system is shown in Fig. 3. In order to recognize each single AU we construct a fuzzy rule-based system. The reduced feature vector is used as the input of the system.

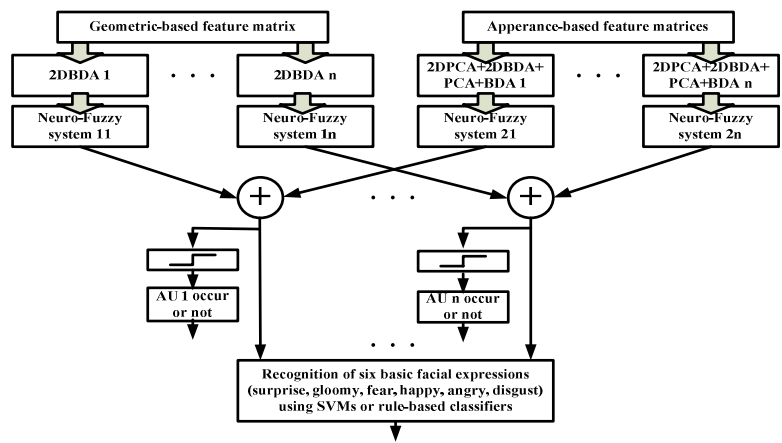

**Fig. 3.** Block diagram of the proposed system (n is number of the facial action units)

Each system is composed of n Takagi-Sugeno type fuzzy if-then rules of the below format:

$R^i$: if $x_1$ is $A_1^i$ and ... and $x_k$ is $A_k^i$ then $y = p_0^i + p_1^i x_1 + \cdots + p_k^i x_k$ \hspace{1em} $(i = 1, \ldots, n)$

Here, y is variable of the consequence whose value is the AU intensity and we should infer it. $x_1, \ldots, x_k$ are variables of the premise, i.e. features, that appear also in the part of the consequence. $A_1^i, \ldots, A_k^i (i = 1, \ldots, n)$ are fuzzy sets representing a fuzzy subspace in which the rule $R^i$ can be applied for reasoning (we use Gaussian membership function with two parameters), and $p_0^i, p_1^i \ldots, p_k^i$ $(i = 1, \ldots, n)$ are consequence parameters. The fuzzy implication is based on a fuzzy partition of the input space. In each fuzzy subspace, a linear input-output relation is formed.

When we are given $(x_1 = x_1^0, x_2 = x_2^0, \ldots, x_k = x_k^0)$, the fuzzy inference system produces output of the system as follows:

For each implication $R^i$, $y_i$ is calculated as:

$$y_i = p_0^i + p_1^i x_1^0 + \cdots + p_k^i x_k^0 \hspace{2em} i = 1, \ldots, n \hspace{2em} (3)$$

The weight of each proposition $y = y_i$ is calculated as:

$$w_i = A_1^i(x_1^0) \times \ldots \times A_k^i(x_k^0) \hspace{2em} i = 1, \ldots, n \hspace{2em} (4)$$

Then, the final output y inferred from n rules is given as the average of all $y_i$ ($i = 1, \ldots, n$) with the weights $w_i$, i.e. $y = (\sum_{i=1}^{n} y_i w_i)/(\sum_{i=1}^{n} w_i)$.

For modeling each system, i.e. each single AU, we should extract the rules using training data. Depending on the AU that we want to model it, the sequences in the training set are divided into two subsets: positive set and negative set. The positive set includes all of sequences that the AU occurs on them; other sequences are placed in the negative set. Then, by applying the method discussed in section 2, we extract the feature vector for each sequence. Assuming the last frames of the sequences in the training set have maximum intensity, the target value of feature vectors which are in the positive set and negative set is labeled 1 and 0 respectively. We also use some sequences several times with different intensities, i.e. by using intermediate frames as the last frame and removing the frames which come after it. For these sequences, if the original sequence was negative the target value is again 0. Otherwise, the target value of corresponding feature vector for produced sequence is calculated by:

$$t = \frac{\text{sumdistances}_1}{\text{sumdistances}_2} \hspace{2em} (5)$$

where $\text{sumdistances}_1$ is the sum of the Euclidian distances between point of the Wincandide-3 grid in the last frame of the produced sequence and their positions in the first frame (a subset of points for upper face and lower face action units are used). Similarly, $\text{sumdistances}_2$ is the sum of the Euclidean distances between points of the model in the last frame of the original sequence and their positions in the first frame; e.g. if we remove all frames which come after the first frame then $t = 0$, and if we remove none of the frames then $t = 1$. We model each single AU two times, using geometric and appearance features separately. In test phase, the outputs are added and the result is passed through a threshold. When several outputs were on, it signals that a combination of AUs has been occurred.

Modeling of each system, i.e. each single AU, is composed of two parts: structure identification and parameter identification. The structure identification relate to partition of the input space, i.e. number of rules. In parameter identification process the premise parameters and consequence parameters are determined.

### 3.2 Structure Identification Algorithm

We use some of the training samples as the validation set. This set is used for avoiding the problem of overfitting data in the modeling process. Suppose $x_1, x_2, \ldots, x_k$ are the inputs of the system. Moreover, suppose $d_i$ is the number of divided fuzzy subspace for $x_i$. The initial value of $d_i$ ($i = 1, \ldots, k$) is 1, because at first the range of each variable is undivided. Also, let V, i.e. the value of mean squares of errors of the model on validation set be a big number. The algorithm of modeling is as follows:

1) The range of $x_1$ is divided into one more fuzzy subspace (e.g. "big" and "small" if $d_1 = 1$ or "big", "medium" and "small" if $d_1 = 2$) and the range of the other variables $x_2, x_3, \ldots, x_k$ are not more divided. This model is called model 1; e.g. in the first iteration, model 1 consisting of two rules of the below format:

$$R^1: \text{if } x_1 \text{ is big then } y = p_0^1 + p_1^1 x_1$$

$$R^2: \text{if } x_1 \text{ is samll then } y = p_0^2 + p_1^2 x_1$$

Similarly the model in which the range of $x_2$ is divided in to one more subspace and the rage of other variables $x_1, x_3, \ldots, x_k$ are not more divided, is called mode 2. In this way we have k models.

2) For each model the optimum premise parameters (mean and variance of the membership functions) and consequence parameters are found by the parameter identification algorithm described in the next subsection. This algorithm applies hybrid learning, to determine the premise and consequence parameters.

3) For each model, the mean squares of errors (MSE) using training data is calculated:

$$\text{MSE} = \frac{\sum_{j=1}^{P}(y^j - t^j)^2}{P} \tag{6}$$

Here, P is number of the training data, $y^j$ ($j = 1, \ldots, P$) is the final output inferred from rules of the model for j'th feature vector in the training set. $t^j$ ($j = 1, \ldots, P$) is target value for j'th input vector in training set, which is a number between 0 to 1. Then, the model with least mean squares of errors is selected. This model called stable state model. Let T be the MSE of the stable state model using validation set.

4) If $T \geq V$ stop otherwise, let $d_s = d_s + 1$. where s is index of the stable state model; let $V = T$ and go to step 1.

After each iteration of the modeling algorithm, the range of a variable is divided to one more fuzzy subspace. In each fuzzy subspace, a linear input-output relation in consequence part of the corresponding rule is used to approximate the intensity of AU. Consequently, a highly non-linear system can be approximated efficiently by this method. Applying this accurate approach, we can recognize the lower intensity and combinations of AUs.

### 3.3 Parameter Identification Using Hybrid Learning

The goal of this section is determining the optimum premise parameters (mean and variance of the membership functions), and consequent parameters of the model, assuming fixed structure. We use an adaptive-network-based fuzzy inference system (ANFIS) to determine the parameters (for more details see [16]). This architecture represents the fuzzy inference described in subsection 4.1. Given the values of premise parameters, the overall output can be express as a linear combinations of consequence parameters. In forward pass of the hybrid learning algorithm, functional signals go forward till layer 4 of the ANFIS and the consequence parameters are identified by the least squares estimate. In the backward pass, the error rates propagate backward and the premise parameters are updated by gradient descent procedure.

Alternatively, we could apply the gradient descent procedure to identify all parameters. But this method is generally slow and likely to become trapped in local minima. By using the hybrid algorithm, we can decrease the dimension of search space and cut down the convergence time.

## 4   Facial Expression Recognition and Interpretation

Although we can use a SVMs for classification of six basic facial expressions (by feature vectors directly or AU intensity values), employing rule-based classifiers such as J48 [17], we can automatically extract human interpretable classification rules to interpret each expression. Thus, novel accurate AU-to-expression converters by continues values of the AU intensities can be created. These converters would be useful in animation, cognitive science, and behavioral science areas.

## 5   Experimental Results

To evaluate the performance of the proposed system and other methods like support vector machines (SVMs), hidden Markov models (HMMs), and neural network (NN) classifiers, we test them on Cohn-Kanade database [18]. The database includes 490 frontal view image sequences from over 97 subjects. The final frame of each image sequence has been coded using Facial Action Coding System which describes subject's expression in terms of action units. For action units that vary in intensity, a 5-point ordinal scale has been used to measure the degree of muscle contraction. In order to test the algorithm in lower intensity situation, we used each sequence five times with different intensities, i.e. by using intermediate frames as the last frame. Of theses, 1500 sequences were used as the training set. Also, for upper face and lower face AUs, 240 and 280 sequences were used as the test set respectively. None of the test subjects appeared in training data set. Some of the sequences contained limited head motion.

Image sequences from neutral to the frame with maximum intensity, were cropped into $57 \times 102$ and $52 \times 157$ pixel arrays for lower face and upper face action units respectively. To extract appearance features we applied 16 Gabor kernels to each frame. We used the same dimension reduction method in the proposed and SVMs methods. Depending on the single AU that we want to model it, the geometric and appearance feature vectors were of dimension 4 to 8 after applying the dimension reduction techniques. In training phase we allowed the target value of feature vector for multiple systems (single AUs) to set 1, when the input consists of AU combinations. The value of the threshold is set to 1 (see Fig. 3).

Table 1 and Table 2 show the upper face and lower face action unit recognition results respectively. In the proposed method, an average recognition rate of 88.8 and 95.4 percent were achieved for upper face and lower face action units respectively. Also, an average false alarm rate of 7.1 and 2.9 percent were achieved for upper face and lower face action units respectively. In SVMs method, we first concatenated the reduced geometric and appearance feature vectors for each single AU. Then, we classify them using a two-class SMVs classifier with Gaussian kernel. Due to use of crisp value for targets, this method suffers from intensity variations. In HMMs method, the best performance was obtained by three Gaussians and five states. The Gabor coefficients were reduced to 100 dimensions per image sequence using PCA and 2DPCA (like [19]). The geometric features were reduced to 8 dimension using PCA. Then, we concatenated the geometric and appearance feature vectors. For each single AU and also each AU combination, a hidden Markov model was trained, i.e. in this method we consider each AU as a class. We used the same dimension reduction method in the NN and HMMs methods.

**Table 1.** Upper face action unit recognition results (R=recognition rate, F=false alarm)

| Proposed method | | | | |
|---|---|---|---|---|
| AUs | Sequences | Recognized AUs | | |
| | | True | Missing or extra | False |
| 1 | 20 | 17 | 2(1+2+4), 1(1+2) | 0 |
| 2 | 10 | 7 | 1(1+2+4), 2(1+2) | 0 |
| 4 | 20 | 19 | 1(1+2+4) | 0 |
| 5 | 20 | 20 | 0 | 0 |
| 6 | 20 | 19 | 0 | 1(7) |
| 7 | 10 | 9 | 0 | 1(6) |
| 1+2 | 40 | 37 | 2(2), 1(1+2+4) | 0 |
| 1+2+4 | 20 | 18 | 1(1), 1(2) | 0 |
| 1+2+5 | 10 | 8 | 2(1+2) | 0 |
| 1+4 | 10 | 7 | 3(1+2+4) | 0 |
| 1+6 | 10 | 8 | 1(1+6+7) | 1(7) |
| 4+5 | 20 | 17 | 2(4), 1(5) | 0 |
| 6+7 | 30 | 27 | 2(1+6+7), 1(7) | 0 |
| Total | 240 | 213 | 24 | 3 |
| R | 88.8% | | | |
| F | 7.1% | | | |

| HMMs | | | | |
|---|---|---|---|---|
| AUs | Sequences | Recognized AUs | | |
| | | True | Missing or extra | False |
| 1 | 20 | 15 | 2(1+2+4), 2(1+2) | 1(2) |
| 2 | 10 | 6 | 2(1+2+4), 1(1+2) | 1(1) |
| 4 | 20 | 18 | 1(1+2+4) | 1(2) |
| 5 | 20 | 20 | 0 | 0 |
| 6 | 20 | 18 | 1(1+6) | 1(7) |
| 7 | 10 | 7 | 2(6+7) | 1(6) |
| 1+2 | 40 | 38 | 1(1+2+4) | 1(4) |
| 1+2+4 | 20 | 16 | 2(2), 2(1+2) | 0 |
| 1+2+5 | 10 | 7 | 3(1+2) | 0 |
| 1+4 | 10 | 5 | 3(1+2+4) | 2(5) |
| 1+6 | 10 | 6 | 2(1+6+7) | 2(7) |
| 4+5 | 20 | 14 | 3(4), 1(5) | 2(2) |
| 6+7 | 30 | 25 | 2(1+6+7), 3(7) | 0 |
| Total | 240 | 195 | 33 | 12 |
| R | 81.3% | | | |
| F | 12.9% | | | |

| SVMs | | | | |
|---|---|---|---|---|
| AUs | Sequences | Recognized AUs | | |
| | | True | Missing or Extra | False |
| 1 | 20 | 15 | 2(1+2+4), 1(1+2) | 2(2) |
| 2 | 10 | 6 | 2(1+2+4) | 2(1) |
| 4 | 20 | 18 | 1(1+2+4) | 1(2) |
| 5 | 20 | 20 | 0 | 0 |
| 6 | 20 | 19 | 1(1+6) | 0 |
| 7 | 10 | 7 | 0 | 3(6) |
| 1+2 | 40 | 35 | 1(2), 2(1+2+4) | 2(4) |
| 1+2+4 | 20 | 15 | 2(1), 2(2) | 1(5) |
| 1+2+5 | 10 | 6 | 2(1+5) | 2(4) |
| 1+4 | 10 | 4 | 3(1+2+4) | 3(5) |
| 1+6 | 10 | 6 | 3(1+6+7) | 1(7) |
| 4+5 | 20 | 15 | 2(1+2+5) | 3(2) |
| 6+7 | 30 | 24 | 2(1+6+7), 2(7) | 2(1) |
| Total | 240 | 190 | 28 | 22 |
| R | 79.2% | | | |
| F | 17.1% | | | |

| NN | | | | |
|---|---|---|---|---|
| AUs | Sequences | Recognized AUs | | |
| | | True | Missing or Extra | False |
| 1 | 20 | 14 | 3(1+2+4) | 3(2) |
| 2 | 10 | 5 | 4(1+2+4) | 1(1) |
| 4 | 20 | 18 | 1(1+2+4) | 1(2) |
| 5 | 20 | 18 | 1(4+5) | 1(5) |
| 6 | 20 | 18 | 2(1+6) | 0 |
| 7 | 10 | 6 | 2(6+7) | 2(6) |
| 1+2 | 40 | 36 | 2(2), 2(1+2+4) | 0 |
| 1+2+4 | 20 | 16 | 2(1), 2(2) | 0 |
| 1+2+5 | 10 | 7 | 2(1+2) | 1(4) |
| 1+4 | 10 | 5 | 4(1+2+4) | 1(5) |
| 1+6 | 10 | 6 | 2(1+6+7) | 2(7) |
| 4+5 | 20 | 16 | 3(4) | 1(1) |
| 6+7 | 30 | 24 | 3(1+6+7), 3(7) | 0 |
| Total | 240 | 189 | 38 | 13 |
| R | 78.8% | | | |
| F | 15.4% | | | |

Although this method can deal with AU dynamics properly, it needs the probability density function for each state. Moreover, the number of AU combinations is too big and the density estimation methods may lead to poor result especially when the number of training samples is low. Finally, in NN methods we trained a neural network with an output unit for each single AU and by allowing multiple output units to fire when the input sequence consists of AU combinations (like [20]).

**Table 2.** Lower facial action unit recognition results (R=recognition rate, F=false alarm)

| Proposed method | | | | |
|---|---|---|---|---|
| AUs | Sequences | Recognized AUs | | |
| | | True | Missing or extra | False |
| 9 | 8 | 8 | 0 | 0 |
| 10 | 12 | 12 | 0 | 0 |
| 12 | 12 | 12 | 0 | 0 |
| 15 | 8 | 8 | 0 | 0 |
| 17 | 16 | 16 | 0 | 0 |
| 20 | 12 | 12 | 0 | 0 |
| 25 | 48 | 48 | 0 | 0 |
| 26 | 24 | 18 | 4(25+26) | 2(25) |
| 27 | 24 | 24 | 0 | 0 |
| 9+17 | 24 | 24 | 0 | 0 |
| 9+17+23+24 | 4 | 3 | 1(19+17+24) | 0 |
| 9+25 | 4 | 4 | 0 | 0 |
| 10+17 | 8 | 5 | 2(17), 1(10) | 0 |
| 10+15+17 | 4 | 3 | 1(15+17) | 0 |
| 10+25 | 8 | 8 | 0 | 0 |
| 12+25 | 16 | 16 | 0 | 0 |
| 12+26 | 8 | 6 | 2(12+25) | 0 |
| 15+17 | 16 | 16 | 0 | 0 |
| 17+23+24 | 8 | 8 | 0 | 0 |
| 20+25 | 16 | 16 | 0 | 0 |
| Total | 280 | 267 | 11 | 2 |
| R | 95.4% | | | |
| F | 2.9% | | | |

| HMMs | | | | |
|---|---|---|---|---|
| AUs | Sequences | Recognized AUs | | |
| | | True | Missing or extra | False |
| 9 | 8 | 8 | 0 | 0 |
| 10 | 12 | 11 | 1(10+17) | 0 |
| 12 | 12 | 12 | 0 | 0 |
| 15 | 8 | 6 | 1(15+17) | 1(17) |
| 17 | 16 | 16 | 0 | 0 |
| 20 | 12 | 12 | 0 | 0 |
| 25 | 48 | 45 | 2(25+26) | 1(26) |
| 26 | 24 | 19 | 3(25+26) | 2(25) |
| 27 | 24 | 24 | 0 | 0 |
| 9+17 | 24 | 22 | 2(9) | 0 |
| 9+17+23+24 | 4 | 2 | 2(19+17+24) | 0 |
| 9+25 | 4 | 4 | 0 | 0 |
| 10+17 | 8 | 3 | 3(10+12) | 2(12) |
| 10+15+17 | 4 | 2 | 2(15+17) | 0 |
| 10+25 | 8 | 7 | 1(25) | 0 |
| 12+25 | 16 | 16 | 0 | 0 |
| 12+26 | 8 | 5 | 2(12+25) | 1(25) |
| 15+17 | 16 | 16 | 0 | 0 |
| 17+23+24 | 8 | 6 | 1(17+23) | 1(10) |
| 20+25 | 16 | 12 | 2(20+26) | 2(26) |
| Total | 280 | 248 | 22 | 10 |
| R | 88.6% | | | |
| F | 8.9% | | | |

| SVMs | | | | |
|---|---|---|---|---|
| AUs | Sequences | Recognized AUs | | |
| | | True | Missing or extra | False |
| 9 | 8 | 8 | 0 | 0 |
| 10 | 12 | 8 | 2(10+7) | 2(17) |
| 12 | 12 | 12 | 0 | 0 |
| 15 | 8 | 6 | 2(15+17) | 0 |
| 17 | 16 | 14 | 2(10+17) | 0 |
| 20 | 12 | 12 | 0 | 0 |
| 25 | 48 | 43 | 2(25+26) | 3(26) |
| 26 | 24 | 18 | 3(25+26) | 3(25) |
| 27 | 24 | 24 | 0 | 0 |
| 9+17 | 24 | 22 | 2(9) | 0 |
| 9+17+23+24 | 4 | 1 | 3(9+17+24) | 0 |
| 9+25 | 4 | 4 | 0 | 0 |
| 10+17 | 8 | 2 | 4(10+12) | 2(12) |
| 10+15+17 | 4 | 2 | 2(15+17) | 0 |
| 10+25 | 8 | 7 | 1(25) | 0 |
| 12+25 | 16 | 16 | 0 | 0 |
| 12+26 | 8 | 3 | 3(12+25) | 2(25) |
| 15+17 | 16 | 16 | 0 | 0 |
| 17+23+24 | 8 | 6 | 2(17+24) | 0 |
| 20+25 | 16 | 11 | 3(20+26) | 2(26) |
| Total | 280 | 235 | 31 | 14 |
| R | 83.9% | | | |
| F | 12.5% | | | |

| NN | | | | |
|---|---|---|---|---|
| AUs | Sequences | Recognized AUs | | |
| | | True | Missing or extra | False |
| 9 | 8 | 7 | 1(9+17) | 0 |
| 10 | 12 | 8 | 2(10+7) | 2(17) |
| 12 | 12 | 11 | 1(12+25) | 0 |
| 15 | 8 | 6 | 2(15+17) | 0 |
| 17 | 16 | 13 | 2(10+17) | 1(10) |
| 20 | 12 | 12 | 0 | 0 |
| 25 | 48 | 42 | 3(25+26) | 3(26) |
| 26 | 24 | 19 | 2(25+26) | 3(25) |
| 27 | 24 | 23 | 1(27+25) | 0 |
| 9+17 | 24 | 22 | 2(9) | 0 |
| 9+17+23+24 | 4 | 1 | 3(9+17+24) | 0 |
| 9+25 | 4 | 4 | 0 | 0 |
| 10+17 | 8 | 4 | 2(10+12) | 2(12) |
| 10+15+17 | 4 | 2 | 2(15+17) | 0 |
| 10+25 | 8 | 7 | 1(25) | 0 |
| 12+25 | 16 | 16 | 0 | 0 |
| 12+26 | 8 | 3 | 3(12+25) | 2(25) |
| 15+17 | 16 | 16 | 0 | 0 |
| 17+23+24 | 8 | 6 | 2(17+24) | 0 |
| 20+25 | 16 | 12 | 3(20+26) | 1(26) |
| Total | 280 | 234 | 32 | 14 |
| R | 83.6% | | | |
| F | 12.9% | | | |

The best performance was obtained by one hidden layer. Although this method can deal with intensity variations by using continues values for target of feature vectors, it suffer from trapping in local minima. Also unlike the proposed method, in NN classifier there is no any systematic approach to determine the structure of the network, i.e. number of hidden layer and hidden units. Table 3 shows the facial expression recognition results using J48 [17] classifier. As discussed in section 4, by applying each rule-based classifier we can develop an AU-to-expression converter.

**Table 3.** Facial expression recognition results using J48 [17] classifier

| Confusion matrix for J48 classifier (total number of samples=2916, correctly classified samples=2710 (92.935%), incorrectly classified samples=206 (7.065%) : | | | | | | The resulted tree for converting the AU intensities to expressions using J48. Each path from root to leaf represents a rule (S=surprise, G=gloomy, F=fear, H=happy, A=angry, D=disgust, the value of each AU is between 0 and 1): |
|---|---|---|---|---|---|---|
| Classified as → | Surprise | Gloomy | Fear | Happy | Angry | Disgust |
| Surprise | 579 | 7 | 7 | 0 | 7 | 0 |
| Gloomy | 6 | 467 | 0 | 0 | 13 | 0 |
| Fear | 23 | 0 | 402 | 0 | 49 | 0 |
| Happy | 0 | 0 | 0 | 618 | 0 | 0 |
| Angry | 29 | 11 | 0 | 0 | 404 | 18 |
| Disgust | 6 | 0 | 0 | 0 | 30 | 24 |
| Detailed accuracy by class for J48 classifier: | | | | | | |
| True positive rate | False positive rate | Precision | ROC area | Class | | |
| 0.965 | 0.028 | 0.900 | 0.992 | Surprise | | |
| 0.961 | 0.007 | 0.963 | 0.996 | Gloomy | | |
| 0.848 | 0.003 | 0.983 | 0.987 | Fear | | |
| 1.000 | 0.000 | 1.000 | 1.000 | Happy | | |
| 0.874 | 0.040 | 0.803 | 0.973 | Angry | | |
| 0.870 | 0.007 | 0.930 | 0.987 | Disgust | | |

Decision tree:
```
AU12 <= 0
|   AU20 <= 0
|   |   AU9 <= 0
|   |   |   AU15 <= 0
|   |   |   |   AU27 <= 0
|   |   |   |   |   AU26 <= 0.083173
|   |   |   |   |   |   AU1 <= 0.268941
|   |   |   |   |   |   |   AU24 <= 0
|   |   |   |   |   |   |   |   AU7 <= 0
|   |   |   |   |   |   |   |   |   AU4 <= 0.152609: A
|   |   |   |   |   |   |   |   |   AU4 > 0.152609: G
|   |   |   |   |   |   |   |   AU7 > 0: A
|   |   |   |   |   |   |   AU24 > 0: A
|   |   |   |   |   |   AU1 > 0.268941
|   |   |   |   |   |   |   AU5 <= 0.360907
|   |   |   |   |   |   |   |   AU16 <= 0: G
|   |   |   |   |   |   |   |   AU16 > 0: F
|   |   |   |   |   |   |   AU5 > 0.360907: S
|   |   |   |   |   AU26 > 0.083173: S
|   |   |   |   AU27 > 0: S
|   |   |   AU15 > 0
|   |   |   |   AU2 <= 0.838493
|   |   |   |   |   AU24 <= 0.935031: G
|   |   |   |   |   AU24 > 0.935031
|   |   |   |   |   |   AU4 <= 0.390682: G
|   |   |   |   |   |   AU4 > 0.390682: A
|   |   |   |   AU2 > 0.838493
|   |   |   |   |   AU5 <= 0.360907: G
|   |   |   |   |   AU5 > 0.360907: S
|   |   AU9 > 0
|   |   |   AU24 <= 0: D
|   |   |   AU24 > 0: A
|   AU20 > 0
|   |   AU27 <= 0.360907: F
|   |   AU27 > 0.360907: S
AU12 > 0
|   AU2 <= 0.390682
|   |   AU4 <= 0.268941: H
|   |   AU4 > 0.268941: F
|   AU2 > 0.390682: S
```

## 6 Discussion and Conclusions

We proposed an efficient system for representation, recognition, interpretation, and analysis of the facial action units (AUs) and expressions. As an accurate tool, this system can be applied to many areas such as recognition of spontaneous and deliberate facial expressions, multi modal/media human computer interaction and lie detection efforts. In our neuro-fuzzy classification scheme each fuzzy rule applies a linear approximation to estimate the AU intensity in a specific fuzzy subspace. In addition combining geometric and appearance features increases the recognition rate.

Although the computational cost of the proposed method can be high in the training phase, when the fuzzy inference systems were created, it needs only some matrix products to reduce the dimensionality of the geometric and appearance features in the test phase. Employing a 3× 3 Gabor kernel and a grid with low number of vertices, we can construct the Gabor representation of the input image sequence and also track the grid in less than two seconds with moderate computing power. As a result, the proposed system is suitable for real-time applications. Future research direction is to consider variations on face pose in the tracking algorithm.

**Acknowledgment.** The authors would like to thank the Robotic Institute of Carnegie Mellon University for allowing us to use their database.